\documentclass[sigconf]{acmart}
\usepackage{algorithm2e}
\usepackage{cleveref}
\usepackage{mathtools} 
\newtheoremstyle{sltheorem}
{}                
{3pt}             
{\slshape}        
{}                
{\bfseries}       
{.}               
{3pt}               
{}                
\theoremstyle{sltheorem}
\newtheorem{challenge}{Challenge}
\AfterEndEnvironment{challenge}{\noindent\ignorespaces}

\setcopyright{rightsretained}
\copyrightyear{2021}
\acmYear{2021}
\acmDOI{}

\acmConference[KDD MLF '21]{KDD Workshop on Machine Learning for Finance}{August 15, 2021}{Virtual}
\acmPrice{}
\acmISBN{}

\usepackage{enumitem}
\usepackage{lipsum}
\setlist[itemize]{leftmargin=*}
\begin{document}


\title{Beyond Fairness Metrics: \\
Roadblocks and Challenges for Ethical AI in Practice} 

\author{Jiahao Chen}
\email{jiahao.chen@jpmchase.com}
\orcid{0000-0002-4357-6574}
\affiliation{%
  \institution{J.\ P.\ Morgan AI Research}
  \streetaddress{383 Madison Avenue}
  \city{New York}
  \state{New York}
  \country{USA}
  \postcode{10179-0001}
}
\author{Victor Storchan}
\email{victor.storchan@jpmchase.com}
\affiliation{%
  \institution{J.\ P.\ Morgan}
  \streetaddress{310 University Avenue}
  \city{Palo Alto}
  \state{California}
  \country{USA}
  \postcode{94301-1744}
}

\author{Eren Kurshan}
\email{ek2529@columbia.edu}
\affiliation{%
    \institution{Columbia University}
    \city{New York}
    \state{New York}
    \country{USA}
}


%


\begin{abstract}
We review practical challenges in building and deploying ethical AI
at the scale of contemporary industrial and societal uses.
Apart from the purely technical concerns that are the usual focus of academic research,
the operational challenges of inconsistent regulatory pressures, conflicting business goals,
data quality issues, development processes, systems integration practices,
and the scale of deployment all conspire to create new ethical risks.
Such ethical concerns arising from these practical considerations
are not adequately addressed by existing research results.
We argue that a holistic consideration of ethics in the development and deployment of AI systems
is necessary for building ethical AI in practice,
and exhort researchers to consider the full operational contexts of AI systems
when assessing ethical risks.
\end{abstract}

\settopmatter{printfolios=true} 
\maketitle




\section{Introduction}
\label{sec:intro}
\noindent Over the past decade, AI solutions have been deployed pervasively and at scale
in many application areas of societal importance,
like healthcare, education, transportation, financial services, manufacturing, criminal justice,
energy, government, retail, and defense \citep{MckinseyAIUse}.
The massive scale of AI deployments create enormous concentrations of wealth and power.
For example, Facebook and Google account for a combined 60\% of the US online advertising market,
while Amazon controls a third of the cloud business, 44\% of e-commerce, and a staggering 70\% of the smart-speaker market.
Similarly, Google controls over 90\% of the search market \cite{hbr}.
Such concentrations of scale pose risks for causing and amplifying ethical harms,
even human rights issues \citep{Danks2017},
especially when used for high-stakes decisions \citep{Greene2019}.
The growth of high-stakes AI uses and risks has fuelled the growth of AI ethics as an emerging field of academic study.
Nevertheless, studies in this area often fail to address the long term ethical issues plaguing real world usage of AI \citep{Theodorou2020},
focusing instead on narrow technical criteria like fairness metrics \citep{Mehrabi2019,fairmlbook},
or solipsistic philosophical arguments centered around abstract ethical principles,
which are often neither necessary nor sufficient to realize ethical AI in practice \citep{Mittelstadt2019}.

\noindent In this paper, we argue that studies of AI ethics must necessarily
consider the concrete regulatory, organizational and societal contexts
that drive the use of AI,
and tackle the irreducible complexity of deployed AI systems
not just in terms of data and compute complexity,
but also the pragmatics of practice and regulations that govern the application of
core AI technologies like computer vision, natural language processing, data privacy, and autonomous systems control.
Such practical considerations must be capable of handling inconsistencies in regulatory needs across applications and jurisdictions \citep{Cognilytica},
and be able to foreshadow legal and regulatory needs as they evolve to keep up with bleeding-edge innovation.

\noindent This paper does not aim to provide a comprehensive list of challenges in the practice of ethical AI. Instead, it tries to highlight some of the most prominent issues and challenges that go beyond the sophisticated fairness metrics and guidelines in achieving ethical solutions. The paper is organized into specific challenges that are grouped into several sections:
\Cref{sec:regulatory} discusses the regulatory and legal challenges in the practical implementation of AI solutions; \Cref{sec:business} highlights the conflicting business forces and design objectives companies face; \Cref{sec:model} argues that model development and data quality processes are of critical importance in achieving ethical AI solutions; \Cref{sec:scale} argues that the massive scale of AI based applications fundamentally challenge the assumption about discoverability; \Cref{sec:organization} describes organizational and community challenges that need to solved for ethical AI.
We conclude in \Cref{sec:conclusion} with an opinionated outlook for future research.

\section{Regulatory Challenges}
\label{sec:regulatory}

\begin{challenge}
AI regulation is reactionary and not ubiquitous.
\end{challenge} 

\noindent Ethical AI systems must be legal and compliant with regulations, and also governed to ensure trust \citep{Winfield2018}.
However, regulations generally lag behind the cutting edge of AI innovation.
For example, fair lending laws in the U.S.A. are nearly 50 years old and are only now just starting to be modernized
to handle the use of AI systems in practice \cite{usrfi}.
The experiences of other fields like bioethics show that technological regulation depends heavily on societal values,
and therefore is in a state of continuous reform as perspectives evolve over time.
However, the current pace of regulatory evolution is out of sync with the frenetic pace of developments in AI research and practice;
top AI conferences like NeurIPS report 50\% annual increases in their submissions \citep{Neuripspapers}.
This rapid growth in AI complexity and subsequent opacity, coupled with the intrinsic probabilistic nature of core AI technologies,
leads to challenges in establishing best practices that are both state-of-the-art and compliant with applicable regulations.

\noindent \textit{AI systems have growing scope:}
As noted above, AI systems are used in many industries, but each industry is often regulated in its own siloed domain,
with little regulation governing the use of AI except in specific industries such as healthcare and financial services,
or in specific corporate practices such as employment \citep{Bogen2018,amazon}.
In general, no single regulatory body is responsible for the totality of AI usage \citep{Prates2020},
which creates the risk of overlooking systemic effects due to the composition of multiple AI systems
across multiple applications and organizations.

\noindent \textit{Current regulations:}
In the past decade, regulations like GDPR \citep{GDPR} and CCPA \citep{CCPA} have significantly advanced transparency and explainability requirements.
However, current AI techniques for explainability, fairness and other ethical concerns
are far from fully addressing accountability needs \citep{Goebel2018,Barredo-Arrieta2020}.
For example, the right to non-discrimination in GDPR, while reflecting a fundamental right in the E.U.,
leaves ambiguity in enforcement, as the specific definitions of protected classes and analogous concepts is devolved
to individual countries rather than defining a pan-European standard \citep{Wachter2021b}.
Since ``fairness through unawarness'' has been thoroughly discredited as a meaningful way to enforce AI fairness,
the collection of relevant demographic data is therefore necessary to audit and remediate bias \citep{Goodman2017, Chen2019}.
However, the lack of uniform standards poses a major operational challenge for organizations wishing to scale across the entire E.U.\ market.
Therfore, there is a need for human-intelligible explanations of algorithmic decisions to address multifaceted compliance needs,
as one size does not fit all.

\begin{challenge}
AI regulations have practice enforcement challenges.
\end{challenge} 

\noindent
The inconsistency of regulations across multiple jurisdictions, industries, and use cases poses challenges not just for compliance,
but also for enforcement by regulators.
Many AI-specific regulations lack enforcement power, relying instead on self-regulation, which carries intrinsic risks of conflicts of interest \citep{Resseguier2020}.
Happily, individual programs such as the
European Commission’s research funding schemes now require ethics appraisal procedures \citep{EU_Operationalize},
which will provide valuable lessons for future regulations.
Nevertheless, new complexity challenges arise from the growing scale of data, compute, and systems integrations,
as well as growing demands on data privacy, security and intellectual property protections.
These challenges require new governance mechanisms, not just to handle growing real-time and continuous oversight and enforcement,
but also to reason about subtle holistic compositional effects and trade-offs between conflicting stakeholder needs
\citep{Clark2019,Kurshan2020}.

\begin{challenge}
Gathering evidence for accountability poses risks for remediation.
\end{challenge}

\noindent
The law is a powerful mechanism for holding users of AI systems accountable \citep{Doshi-Velez2017}.
However, collecting the necessary evidence of wrongdoing is often difficult in practice
until the problem becomes too extensive to ignore.
The burden of evidence becomes a particular concern due to the feedback loops inherent in
decisions like predictive policing or credit decisioning \citep{fairmlbook}.
Gathering evidence in itself thus risks ethical harms of inaction while self-fulfilling predictions
amplify discriminatory bias over time, particularly in predictive tools such as affinity profiling
which can cause discrimination outside the legal framework of protected classes \citep{Wachter2021}.
The people most capable of early detection and action are therefore the people
most in need of legal protections like whistleblower laws, which unfortunately exist only in
very limited circumstances such as in the finance industry \citep{Culiberg2017,medium}.

\section{Conflicting Business Forces \& Objectives}
\label{sec:business}

\begin{challenge}
Business needs are often in conflict with ethics and transparency.
\end{challenge} 

\noindent An intimate knowledge of AI systems is often needed for detecting potential ethical risks.
However, companies deploying AI systems often treat their solutions as proprietary black boxes
to be protected as trade secrets \citep{Hagendorff2020},
with assumptions and limitations divulged only on a
need-to-know basis to regulators or customers with specific compliance needs.
Such attitudes are usually justified to maintain a competitive edge,
reduce legal liabilities posed by discovery, and to protect intellectual property.
As such, competing financial incentives, coupled with a general absence of 
pressures for accountable and responsible development practices \textit{ex ante} \citep{strubell2019energy,Raji2020},
and the lack of intrinsic legal liability for the use of AI \citep{Sullivan2018},
can prevent the development of ethical AI \citep{Deloitte_Incentive,Hagendorff2020,GoodAI,Kurshan2020}. 
As a result, we have seen that interest in ethics committees has waned \citep{Accenture_AI_Committee},
with several companies have dismantling their oversight committees entirely \citep{BBC,FT}.
Regulation is therefore a powerful tool to incentivize the building and usage of ethical AI \citep{NYTimes},
as well as encourage due attention to both immediate and long-term ethical consequences of AI systems \citep{Prunkl2020}.



\section{Model Development \& Process Challenges}
\label{sec:model}

\begin{challenge}
The need for more data in AI has fuelled reliance on third-party broker data, which often have data quality issues.
\end{challenge}

\noindent \textbf{\textit{Data quality issues in source data:}}
The ubiquity of low-cost computing resources has driven demand for more data for AI systems,
resulting in a growth in data purchased from third-party broker data
to be used in a broad range of AI applications from recruiting to fintech-based loan decisions.
While the use of such broker data is in general unregulated save in specific industries \citep{Hurley2016},
third-party data sources raise the risk of data quality and privacy issues \citep{Kuempel2016,Bender2021}.
Data brokers amass data from wide-ranging sources like social media, public records, digital footprints, bill payments, transaction data, health indicators, alcohol consumption \citep{Experian,FT}, which raise concerns over data quality, privacy, and risks of discrimination \citep{Martin,Rostow2017}. 
Such quality and privacy risks have been highlighted by the Federal Trade Commission,
citing unethical practices during data acquisition \citep{FTC_Data}
as well as reselling such data to scammers and criminals \citep{FTC_Scam}.
However, consumers lack the statutory rights at the federal level to know what information data brokers have about them,
the quality of such data, as well as which data brokers hold any such information \citep{Kuempel2016},
save in specific situations like financial services through fair lending laws like the Fair Credit Reporting Act.
Thus in general, it is difficult for anyone to know the true extent of their data footprint, let alone control its use.

\noindent\textbf{\textit{AI development issues:}}
A discussion of ethical concerns must also incorporate the need to assess if AI systems are built adequately for their intended purpose.
To give one example,
a recent meta-analysis of over 2,000 papers purporting to use AI to detect COVID-19
revealed extensive methodological flaws and biases, such as mixing adult and pediatric pneumonia data in the control group,
so much so that none of the papers they studied were evaluated to have any potential clinical significance \citep{Roberts2021}.
Such cautionary tales highlight the need for a holistic assessment of ethical concerns alongside
other technical challenges such as under-specification \citep{damour2020underspecification},
overfitting, ignorance of causal effects, and data biases.

\section{Too Big To Audit}
\label{sec:scale}

\begin{challenge}
The scale of industrial AI systems poses challenges for discovering ethical concerns.
\end{challenge} 

\noindent\textbf{\textit{Identifying impacted populations:}}
The massive scale of AI systems in practice create the risk for subtle effects to
cascade into significant ethical harms.
For example, discrimination against smaller groups that lie outside the usual legal and regulatory protections
and escape detection without intentional efforts to discover them \citep{Bogen2018}.
Conversely, AI can learn subtle cues that reinforce existing human biases,
e.g. in using subtle facial expressions for loan decisions, personality trait analysis for recruiting,
or surveillance techniques for online proctoring \citep{FTC_EPIC}. 
Without knowledge of class membership in at-risk subgroups,
statistical methods have fundamental limitations to the estimation of such discrimination \citep{Chen2019,Kallus2021,Awasthi2021},
which has motivated recent progress on unsupervised methods for detecting such
anomalously classified data points \citep{hooker2020characterising,agarwal2020estimating}.
Nevertheless, fundamental learning theory constraints remain on the ability to remediate bias, even if detected \citep{Agrawal2021}.

\noindent\textbf{\textit{Scale and resource challenges for assessment:}}
The enormous volumes of data produced by social media, financial systems and other large-scale industrial applications creates
big data challenges just to assess the potential for ethical harms.
Third-party auditors and independent researchers must surmount the challenges of data privacy, storage, management and governance
to even access the data safely, let alone amass the compute resources needed to analyze the data at scale
or retrain models.
A large academic supercomputing center estimated the effort needed to retrain ImageNet from scratch at a cost of over one million dollars \citep{You2017,You2018},
making such independent studies infeasible without extensive funding support.

\section{Community \& Organizational Challenges}
\label{sec:organization}

\begin{challenge}
Despite the assumption of full autonomy, most AI systems are only components of broader human and business processes.
\end{challenge}
In the real world, models rarely operate in complete autonomy. Rather it is the combination of the human assisted by the model's prediction and potentially empowered by the model's explanations which lead to a particular strategic behavior or final decision \citep{tsirtsis2020decisions}. As an example, decision makers would be likely to consider more social-economic factors in combination with the results of the model. As a consequence, explainable AI is needed both for the operator who directly interacts with the model and for the final customers or impacted individuals who is potentially affected by the semi-automated decision. In the banking sector, as the regulation (FCRA regulation on fair lending in particular) is evolving towards more socially responsible behaviors, it is becoming in practice more and more complex to determine what would be the best set of counterfactual explanations to provide to an individual whose loan application would be denied. Optimizing the utility of the counterfactual explanation for a fixed policy is actually a NP-Hard problem \citep{tsirtsis2020decisions}. Moreover, on the practitioner's side, understanding the underlying mechanisms of cognitive psychology in collaborative decision making is still a work in progress. Recent studies show that explainable AI methods produce over-reliance on the model instead of only providing interpretability \citep{Kaur2020,Bansal2021}. Fairness and explainable guidelines and regulations have to take into account this practical negative externalities.

\begin{challenge}
The need for technical leadership for less-regulated industries and applications.
\end{challenge}

\noindent \textit{\textbf{Industry Leadership:}} Due to the increased reliance on digital solutions an increasing number of unregulated companies use AI systems. As a result AI solutions have large scale impact \citep{Terzi2019}. Due to its efficiency, a small number of AI systems hence may have unprecedented impact on the markets and the society. 
Compared to the regulated counterparts unregulated or less regulated industries exhibit faster development cycles for AI solutions. Novel solutions emerge without regulatory scrutiny to ensure fairness. As an example, healthcare solutions offered outside of the regulated constructs such as mobile applications pose discrimination and misdiagnosis risks \citep{Feathers2021}. Given that even applications that were considered as lower/no risk for discrimination have exhibited serious ethical challenges  \citep{Finley2015}
real-life application of the ethical principles and guidelines in industrial settings becomes essential \citep{Pichai2018}.

\noindent\textit{\textbf{Ethics Committees:}} (i) \textit{Ethics Committee Practices and Its Limits:} In recent years, ethics committees have been proposed and deployed to tackle the AI ethics challenges in industrial settings \citep{Accenture_AI_Committee}. Despite their potential for positive impact, ethics committees have limitations \citep{Blades1967}. In addition to their limited power and oversight in the past few years number of reports have emerged on technology companies over-ruling or dismantling their ethics boards and committees  \citep{Financialtimes2}. Other reports indicate increased secretiveness \citep{verge,FT}. (ii) \textit{Incentive Conflicts:} Regardless of the power and employment of ethical boards, the motivation to make decisions that contradict with the companies profitability may be hard to justify in many cases. (iii) \textit{Committee Specs:} Currently, there is limited or no studies on the trade-offs involved in building and employing effective ethics committees \citep{Accenture_AI_Committee}.

\begin{challenge}
The relationship between academic research and industry is complex.
\end{challenge}
AI ethics  has emerged as a flourishing area of research in the past years. Research studies are essential in identifying challenges and developing novel solutions.  According to \citet{Abdalla2021},
technology companies have established funding relationships with academic institutions. While this has strong positive impacts in fostering AI ethics and dissemination of knowledge out of the research studies, the corresponding relationships need to carefully managed to prevent unwanted incentive mechanisms. Even indirect and non-monetary incentives may require special consideration in highly-sensitive topics such as AI ethics (e.g. placement of Ph.D. students, internships, invited talks and other factors may play significant roles in both tenured and non-tenured academics). The value alignment for ethical AI requires special consideration for universities \citep{AcademicEnterprise,Nestle2016,Berkeley_AIethics}. 

\noindent Technology companies have significant responsibility in leading ethical AI as they have key roles in model development, guideline development as well as academic funding \citep{IBM_NotreDame} as a source of funding for a large portion of AI research (e.g. research funding, scholarships, centers) \citep{Knight2020}.  A study finds that 58\% of faculty at four prominent universities have received grants, fellowships, or other financial support from 10+ tech firms. Similarly, corporate funding of top cognitive computing and AI conferences is very prominent (almost all conferences have some involvement). Hence, technology companies and corporations have great responsibility in the direction and execution of the AI ethics for broad range of societal applications especially in unregulated industries. 

\begin{challenge}
Limitations and challenges of current organizational constructs restrict the outcomes in practice.
\end{challenge}
\textit{\textbf{Decision Making Groups:}} In most companies AI design decisions are made by a small group of decision makers. Despite the efforts to carefully manage and balance this with ethics committees, corporations are aware of the risks the current organizational constructs carry.  Corporations have hierarchical decision-making processes (with very few officials make percentage of the decisions) \citep{CorporateDecision}. Unlike the external perception, the decision makers mostly involve business stakeholders and non-technical business executives with limited involvement of AI ethics professionals. In many cases, the details of the design decisions are on a need-to-know basis and involve a very small group of people. Controlled by the access rights, most development groups and employees may have a narrow-scope and limited picture of the model characteristics. This creates the need to adjust organizational constructs to employ AI ethics organizations beyond the ethics committees to oversee the AI development in its full scope from the beginning to the end of the development process as well as the monitoring of the models in deployment. Recently, organizational structures and have challenges have gained interest \citep{Rakova2021,Benjamins2020,Vaiste2019}, yet changes in practice are pending. Similarly, studies like \citet{Raji2020} propose frameworks to solve the existing challenges in algorithmic accountability within organization constructs. 

\noindent \textit{\textbf{Decision Processes:}} AI ethics driven processes, such as internal AI audit, are currently lacking in the industrial settings for AI modeling. Such processes may provide valuable checks and balances for the internal audit functions of AI development \citep{Raji2020}.  Some research studies have argued for alternative structures for ethical decision making in corporations (outside of AI) \citep{James2000}, though these proposals are currently in early stages. According to the recent studies, ethical decisions have significant divergence across different geographies, populations and segments \cite{Awad2018}. Representing such diversity in the decision making process to represent the societal needs from AI applications is essential. While some studies argue for diversity in the development and decision making organizations \citep{GenderRace, wired}, the representation of the divergence in the variations in the ethical perspectives may require carefully designed and targeted solution design efforts.

\section{Conclusions \& Outlook}
\label{sec:conclusion}
Artificial intelligence is used in a wide range of applications across different industries. Furthermore, it is rapidly evolving with new techniques being proposed on a continuous basis. As a result, AI ethics has become one of the most important design criteria in AI systems. However, so far AI ethics has mostly been narrow in terms of its focus with a heavy emphasis on fairness and quantitative criteria. In this paper, we argue that ultimately industrial and practical applications will be the determining factor in ethical behavior of AI. Despite the abundance of fairness metrics and ethical AI research practical problems act as a serious obstacle in achieving ethical systems in real-life deployments. Industrial systems are plagued with numerous challenges including but not limited to conflicting financial incentives, organizational challenges and the lack of regulatory constructs. Solving such problems is of primary importance in improving the chances of ethical AI systems that benefit the society as a whole.

\noindent In this paper, we have described several challenges that
real-world practitioners of AI face, which originate from multiple sources.
Technical innovations in AI are outpacing our ability to regulate them effectively,
creating risks for introducing actual ethical harms when deploying bleeding-edge AI solutions
for applications involving high-stakes decision making.
Not only do regulations need to keep pace with the technical progress,
but our abilities to enforce regulations and protect whistle blowers must also increase in response.
Furthermore, AI systems are seldom deployed in a fully autonomous context,
but rather are used in complex business processes involving intimate human oversight and manual controls.
The innate complexity of these processes hinders our ability to audit 
and surface any potential ethical concerns.
On top of such complexity, businesses have strong motivations
to preserve secrecy, from the standpoints of
competitive edge, intellectual property and liability concerns.
Even if ethical concerns were to be surfaced,
remediating them is also not easy in practice,
particularly if there are perceived or real impacts to the bottom line.
In practice, an organization's ability to address ethical issues
depends critically on the strength, support and
and willpower to drive the necessary changes,
or otherwise run the risk of ethics-washing or falling risk to
groupthink within the leadership.
Academic research has a vital role to play in
providing alternative viewpoints,
but only when such research is well-grounded in the real-world challenges
like the ones we have outlined above,
while at the same time managing conflicts of interest
such as industry funding
that may compromise the efficacy of such research.

\paragraph{Disclaimer}
This paper was prepared for informational purposes in part by the Artificial Intelligence Research  group of JPMorgan Chase \& Co and its affiliates (``JP Morgan''), and is not a product of the Research Department of JP Morgan. JP Morgan makes no representation and warranty whatsoever and disclaims all liability, for the completeness, accuracy or reliability of the information contained herein.  This document is not intended as investment research or investment advice, or a recommendation, offer or solicitation for the purchase or sale of any security, financial instrument, financial product or service, or to be used in any way for evaluating the merits of participating in any transaction, and shall not constitute a solicitation under any jurisdiction or to any person, if such solicitation under such jurisdiction or to such person would be unlawful. © 2021 JPMorgan Chase \& Co. All rights reserved.

\bibliography{bib}
\bibliographystyle{ACM-Reference-Format}
\end{document}